%% file: main.tex
\documentclass[letterpaper, 10 pt,journal,twoside]{IEEEtran}

\IEEEoverridecommandlockouts                              

\usepackage{graphics} 
\usepackage[colorlinks=true,allcolors=black]{hyperref}
\usepackage{stmaryrd} 
\usepackage{epsfig} 
\usepackage{times} 
\usepackage{amsmath} 
\usepackage{amssymb}  
\usepackage{float}
\usepackage{hyperref}
\usepackage{subcaption}
\usepackage{physics}
\usepackage{tikz}
\usepackage{pgfplots}
\usepackage{balance}
\usepackage{tabu}
\usepackage{multirow,booktabs}
\usepackage[linesnumbered,ruled]{algorithm2e}

\newcommand{\sref}[1]{{Section \ref{#1}}}
\newcommand{\fref}[1]{Fig. \ref{#1}}
\newcommand{\tref}[1]{Table \ref{#1}}

\DeclareMathOperator*{\argmin}{arg\,min}

\begin{document}

\title{
Intensity-SLAM: Intensity Assisted Localization and Mapping for Large Scale Environment}

\author{Han Wang$^{1}$, Chen Wang$^{2}$, and Lihua Xie$^{1}$
\thanks{Manuscript received: October, 14, 2020; Revised December, 10, 2020; Accepted January, 30, 2021.}
\thanks{This paper was recommended for publication by Editor Sven Behnke upon evaluation of the Associate Editor and Reviewers' comments.
This work was supported by Delta-NTU Corporate Laboratory for Cyber-Physical Systems under the National Research Foundation Corporate Lab @ University Scheme.} 
\thanks{$^{1}$Han Wang and Lihua Xie are with the School of Electrical and Electronic Engineering,
Nanyang Technological University, 50 Nanyang Avenue, Singapore 639798.
        {\tt\small e-mail: \{wang.han,elhxie\}@ntu.edu.sg}}
\thanks{$^{2}$Chen Wang is with the Robotics Institute, Carnegie Mellon University, Pittsburgh, PA 15213, USA. {\tt\small e-mail: chenwang@dr.com}}
\thanks{Digital Object Identifier (DOI): see top of this page.}
}

\markboth{IEEE Robotics and Automation Letters. Preprint Version. Accepted  January, 2021}{Wang \MakeLowercase{\textit{et al.}}: Intensity-SLAM: Intensity Assisted Localization and Mapping}  
 
\maketitle

\begin{abstract}

\input{body/abstract.tex}

\end{abstract}

\section{INTRODUCTION}
\input{body/Introduction.tex}

\section{RELATED WORK}\label{sec:related-work}
\input{body/RelatedWork.tex}

\section{METHODOLOGY}\label{sec:methodology}
\input{body/Methodology.tex}

\section{EXPERIMENT EVALUATION}\label{sec:experiment}

\input{body/Experiments.tex}

\section{CONCLUSION}\label{sec:conclusion}
\input{body/Conclusion.tex}

\balance
\bibliographystyle{IEEEtran}
\bibliography{IEEEabrv,references}

\end{document}

%% file: body/abstract.tex
Simultaneous Localization And Mapping (SLAM) is a task to estimate the robot location and to reconstruct the environment based on observation from sensors such as LIght Detection And Ranging (LiDAR) and camera. It is widely used in robotic applications such as autonomous driving and drone delivery. Traditional LiDAR-based SLAM algorithms mainly leverage the geometric features from the scene context, while the intensity information from LiDAR is ignored. Some recent deep-learning-based SLAM algorithms consider intensity features and train the pose estimation network in an end-to-end manner. However, they require significant data collection effort and their generalizability to environments other than the trained one remains unclear. In this paper we introduce intensity features to a SLAM system. And we propose a novel full SLAM framework that leverages both geometry and intensity features. The proposed SLAM involves both intensity-based front-end odometry estimation and intensity-based back-end optimization. Thorough experiments are performed including both outdoor autonomous driving and indoor warehouse robot manipulation. The results show that the proposed method outperforms existing geometric-only LiDAR SLAM methods.

%% file: body/Introduction.tex
\IEEEPARstart{L}{ocalization} is one of the fundamental and essential topics in robotics. With the development of the robotic industry, robot localization has become more challenging in the past decades: from known to unknown environments, from simple to complex environments, from static to dynamic environments \cite{sun2016towards}, and from short-term to long-term localization \cite{stenborg2018long}. Traditionally, fixed anchors (or routers) are set up in a pre-defined area and the robot pose is obtained by using the distances of the robot from multiple anchors, \textit{e.g.}, Ultra-Wide Band (UWB) \cite{liu2017cooperative} and Wi-Fi localization \cite{sweatt2015wifi}. However, these methods rely on external setup and they are mainly used in a small scale environment.
To tackle the limitation of traditional localization methods, Simultaneous Localization And Mapping (SLAM) is introduced to estimate the robot pose from on-board sensors \cite{li2020multi}. It is independent of an external setup and hence becomes promising in robotic applications. Based on the perception system used, SLAM can be further divided into Visual SLAM (V-SLAM) and LiDAR SLAM. Compared to Visual SLAM, LiDAR SLAM is more accurate and robust to environmental changes such as weather and illumination \cite{debeunne2020review}. 
\begin{figure}[t]
\begin{center}
\vspace{8pt}
\includegraphics[width=0.99\linewidth]{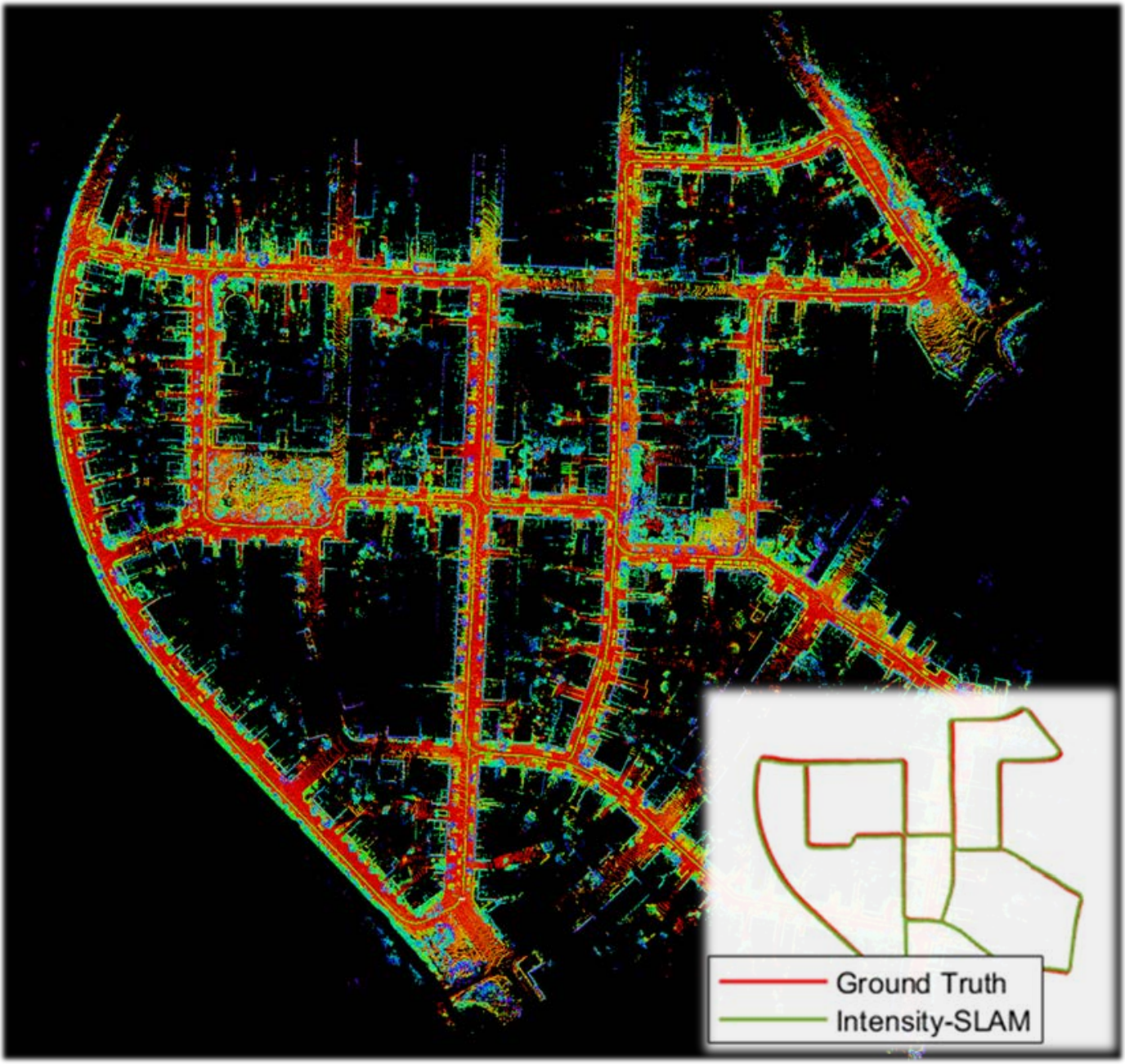}
\captionsetup{justification=justified}
\caption{Example of localization and mapping from the proposed method on KITTI sequence 00. The localization result is shown on the right corner with the ground truth plotted in red color and the trajectory plotted in green color. Our method achieves much higher localization accuracy compared to geometric-only methods. }
\label{fig: title_graph}
\vspace{-0pt}
\end{center}
\end{figure}

Conventional LiDAR SLAM methods such as LOAM \cite{zhang2014loam} and HDL-Graph-SLAM \cite{koide2018portable} mainly focus on the geometric-only information to minimize point cloud differences. 
In this way, the information from the intensity channel is often ignored. Note that the intensity information is related to the reflectivity of material and is different for different objects, which can be useful for localization and object recognition. Therefore, we argue that a robust LiDAR SLAM system should take such intensity information into account. Although recently there are some works aiming to use intensity information to improve the accuracy, they mainly adopt the Convolutional Neural Networks (CNN) to fit the raw laser scan data without any specific intensity analysis or formulation. Moreover, collection of large data is required in order to train neural networks, which can be troublesome in practical applications.

In the previous work \cite{wang2020intensity}, we have demonstrated that intensity information is very useful for robust loop closure detection.
In this work we further utilize the intensity information to improve the localization accuracy of a SLAM system. We propose a novel SLAM framework that uses both geometry and intensity information for odometry estimation. We first analyze the physical model of intensity measurement. Then we introduce an extra intensity cost to the existing geometric-only cost in the odometry estimation formula.
Finally, we combine the intensity-based loop closure detection and back-end optimization to further improve the performance. The proposed framework is tested using an indoor warehouse environment robot as well as an outdoor autonomous driving car. The results indicate that our method provides reliable and accurate localization in multiple environments and outperforms geometric-only methods.
The main contributions of this paper are as below:
\begin{itemize}
\item We propose a novel SLAM framework that uses both intensity and geometry information for localization estimation, which involves front-end odometry estimation and back-end factor graph optimization. The proposed method is open sourced\footnote{\url{https://github.com/wh200720041/intensity_slam}}.
\item We propose to build an intensity map to reveal the intensity distribution and introduce an intensity cost to the existing geometry-only LiDAR SLAM to estimate the robot's location.
\item A thorough evaluation on the proposed method is presented. More specifically, our method is tested in both warehouse and autonomous driving scenarios. The results show that our method outperforms the existing geometric-only methods for multiple scenarios.
\end{itemize}

This paper is organized as follows: \sref{sec:related-work} reviews the related work on existing LiDAR SLAM approaches. \sref{sec:methodology} describes the details of the proposed approach, including intensity calibration, feature selection, odometry estimation, and Intensity Scan Context (ISC) based graph optimization. \sref{sec:experiment} shows the experimental results and comparison with some existing works, followed by the conclusion in \sref{sec:conclusion}.

%% file: body/RelatedWork.tex
Majority of the existing works on LiDAR SLAM focus on geometry information of the environment. One of the most popular methods for point cloud matching is the Iterative Closest Point (ICP) method which matches the current point to the nearest point in the target frame \cite{chetverikov2002trimmed}. ICP iteratively searches for the optimal point correspondence by minimizing the Euclidean distance between point pairs until the transformation matrix converges. The algorithm is popularly used in SLAM systems such as HDL-Graph SLAM \cite{koide2019portable} and LiDAR-Only Odometry and Localization (LOL) \cite{rozenberszki2020lol}. 
However, all points are used for calculation, which is computationally expensive for a LiDAR with tens of thousands of points per scan. ICP is also sensitive to noise. In real applications such as autonomous driving, the measurement noise (\textit{e.g.}, measurement from trees alongside a road) can be significant and subsequently cause localization drifts. 
A more robust and computationally efficient way is to leverage geometric features. In LiDAR Odometry And Mapping (LOAM) \cite{zhang2014loam}, a simple feature extraction strategy is introduced by analyzing the local smoothness. The feature points are segmented into edge features and planar features based on local smoothness, and the robot pose is subsequently calculated by doing edge-to-edge and planar-to-planar matching respectively. 
Similar idea is also used in Lightweight and Ground-Optimized LiDAR Odometry And Mapping (LeGO-LOAM) \cite{shan2018lego}, which targets at providing a computationally efficient LiDAR SLAM for Unmanned Ground Vehicles (UGVs). A laser scan is first segmented into ground points and non-ground points. The planar features are extracted from ground points while the edge features are extracted from non-ground points. The planar features are used to estimate the roll and pitch angle as well as z-translation. The results are subsequently used in the matching of non-ground points to calculate x and y translation as well as yaw angle. LeGO-LOAM has achieved higher localization accuracy than LOAM for grounded robots. 
However, the intensity channel is ignored and only the geometry channel is used in those methods.

In recent years, some works attempt to introduce the intensity channel into SLAM via deep learning. In DeepICP, an end-to-end learning-based 3D point cloud registration is proposed to find the robot pose \cite{lu2019deepicp}. Both intensity channel and geometry channel are introduced into a Deep Feature Extraction (DFE) layer to find the keypoints. Rather than searching for the closest point, a Corresponding Points Generation (CPG) layer is used to generate keypoint correspondences based on learned matching probabilities among a group of candidates, which makes back-propagation possible. Deep-ICP is validated using both KITTI \cite{Geiger2012CVPR} and Apollo-SouthBay dataset \cite{lu2019l3} and outperforms existing methods such as Generalized-ICP \cite{segal2009generalized} and Normal Distribution Transform (NDT) \cite{stoyanov2012fast}. 
Chen \textit{et al} also introduced an end-to-end CNN framework to identify the overlap of two laser scans for back-end SLAM \cite{chen2020rss}. The geometry channel as well as intensity channel are trained through a 11-layer network to generate a 3-dimensional feature. The overlap is estimated by applying another smaller layer to the extracted features, and the yaw angle change is estimated by the cross correlation of the trained features. The experiment on the KITTI dataset shows that the yaw angle estimated from both intensity and geometry channels is better than the geometric-only deep-learning-based framework. 
The deep-learning-based method adopts end-to-end training without further analysis of intensity information. However, it is often difficult and time consuming to collect, label and train data in practice. Moreover, the performance under environment changes may not be consistent, \textit{i.e.}, the trained network may perform well in the environment similar to the training data, but fails when transferring to another environment. 

\begin{figure*}[t]
\begin{center}
\vspace{8pt}
\includegraphics[width=0.95\linewidth]{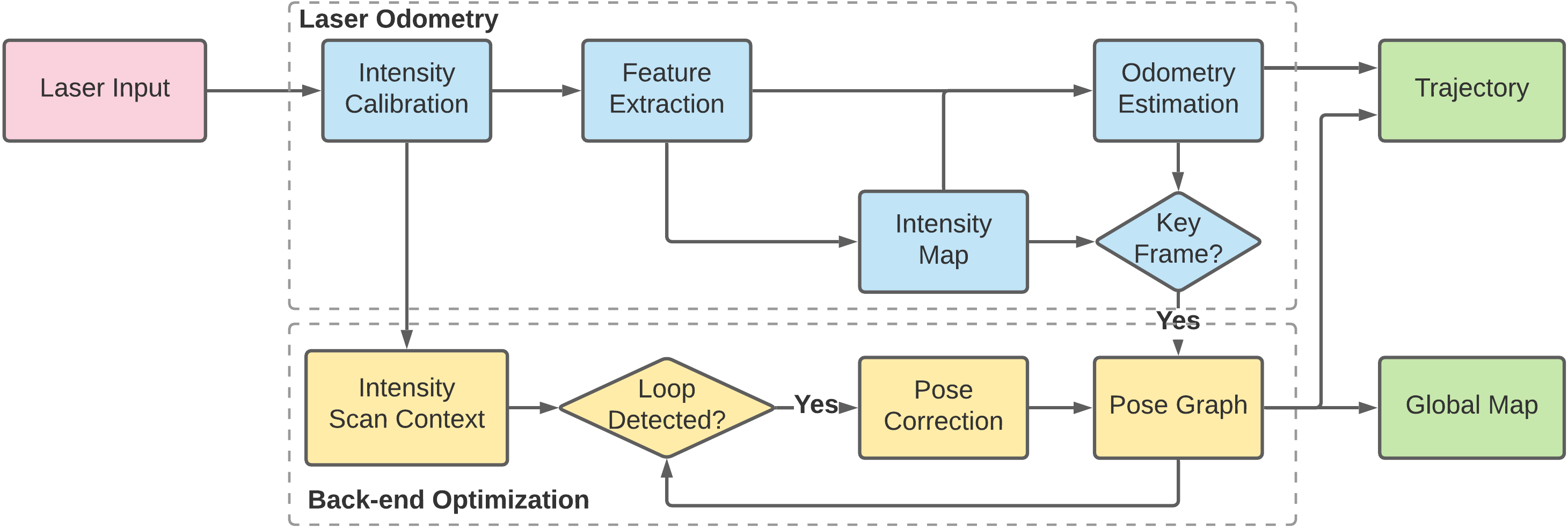}
\captionsetup{justification=justified}
\caption{System overview of proposed Intensity-SLAM. In the front-end, we analyze both geometry and intensity features to estimate odometry. In the back-end, we use intensity scan context for loop closure detection and global pose graph optimization.}
\label{fig: flowchart}
\end{center}
\end{figure*}

Some other works attempt to use the intensity channel and integrate intensity into pose estimation without end-to-end training. Tian \textit{et al} propose an intensity-assisted ICP for fast registration of 2D-LiDAR \cite{tian2019intensity}. Compared to the traditional ICP algorithm, a target function is introduced to determine the initial rigid-body transformation estimation based on the spatial distance and intensity residual. The number of iterations is reduced by 10 times so that the proposed algorithm is able to operate in real time with a single core Central Processing Unit (CPU). In \cite{hewitt2015towards}, an intensity model is incorporated into the Sparse Bundle Adjustment (SBA) \cite{frost2016object} estimation problem. The authors analyze the physical model of intensity measurement and propose a new measurement model that augments 3D localization, intensity and surface normal. Simulation results show that the addition of intensity measurements achieves similar SLAM accuracy with fewer landmarks. This work is further extended in \cite{hewitt2018intense} with the experiment performed using a Time of Flight (ToF) camera. The key-points are selected and tracked via Scale Invariant Feature Transform (SIFT) \cite{lowe1999object} or Binary Robust Invariant Scalable Keypoints (BRISK) \cite{leutenegger2011brisk} feature extraction from an intensity imaginary. The experiment mainly focuses on Visual Odometry (VO) via computer vision techniques but the performance of LiDAR SLAM in large scale environments is not demonstrated. Khan \textit{et al} propose a data driven approach to calibrate the intensity information \cite{khan2016modeling}. Then the scan matching is solved by minimizing the intensity residual of the point pairs rather than geometry residual. It is subsequently integrated into Hector SLAM \cite{KohlbrecherMeyerStrykKlingaufFlexibleSlamSystem2011}, which significantly reduces the drifting error compared to the original Hector SLAM. 
However, it is limited to 2D localization and mapping in a small scale environment.

%% file: body/Methodology.tex

The proposed framework of Intensity-SLAM consists of both front-end laser odometry and back-end loop closure detection. The system overview is shown in \fref{fig: flowchart}. The front-end includes intensity calibration, feature extraction and odometry estimation, while the back-end includes loop closure detection and pose graph optimization. 

\subsection{Intensity Calibration}
LiDAR emits laser beams and measures the arrival time as well as the energy of reflected signals. The object's position in sensor coordinates is determined by emitting angle and distance (\textit{i.e.}, arrival time). The intensity value is determined by the ratio between the energy received from the reflected signals and the laser power emitted. The physical principles of the received power $P_r$ can be determined as \cite{hewitt2015towards}: 
\begin{equation}
  P_r = \frac{P_eD_r^2\rho}{4R^2} {\eta}_{sys} {\eta}_{atm} \cos\alpha,
  \label{equ:phisical model}
\end{equation}
where $P_e$ is the power of emitted laser beam, $D_r$ is the receiver aperture diameter, ${\eta}_{sys}$ is the system transmission factor, ${\eta}_{atm}$ is the atmospheric transmission factor, $\alpha$ is the incident angle between the object surface and laser beam, and $\rho$ is the material reflectivity of the object.
Specifically, the distance measurement $R$ and the incident angle $\alpha$ are extrinsic parameters, while ${\eta}_{sys}$ and ${\eta}_{atm}$ are constant parameters. Therefore, the intensity measurement $I$ is determined by \cite{hewitt2015towards}:
\begin{equation}
\begin{aligned}
 I &= \frac{P_r}{P_e}= \frac{D_r^2\rho}{4R^2}{\eta}_{sys}{\eta}_{atm}\cos\alpha \\
    &= \eta_{all} \frac{\rho \cos \alpha}{R^2},
\end{aligned}
\end{equation}
where $\eta_{all}$ is a constant. Therefore, the surface reflectivity $\rho$ is only related to incident angle $\alpha$ and measured distance $R$ by 
\begin{equation}
  \rho \propto \frac{I R^2}{\cos\alpha}
\end{equation}
For a LiDAR scan, the distance can be easily measured. Hence the incident angle can be estimated by analyzing the local normal. For each point $\textbf{p} \in \mathcal{R}^3$ we can search for the nearest two points $\textbf{p}_1$ and $\textbf{p}_2$, so that the local surface normal $\textbf{n}$ can be expressed as: 
\begin{equation}
 \textbf{n} = \frac{(\textbf{p}-\textbf{p}_1)\times(\textbf{p}-\textbf{p}_2)}{|\textbf{p}-\textbf{p}_1|\cdot|\textbf{p}-\textbf{p}_2|},
\end{equation}
where $\times$ is the cross product. The incident angle is
\begin{equation}
 \cos \alpha = \frac{\textbf{p}^T\cdot\textbf{n}}{|\textbf{p}|}.
\end{equation}
Therefore, we can derive the calibrated intensity scan $\Tilde{\mathcal{I}}$ from a raw LiDAR scan with geometric reading $\mathcal{P}$ and intensity reading $\mathcal{I}$. Specifically, the intensity is partially calibrated with distance measurement by default in some sensors, hence we only apply the calibration to the incident angle. Moreover, low intensity value often leads to a lower Signal-to-Noise Ratio (SNR), which also deteriorates the ranging accuracy for intensity measurement with low value \cite{lin2020loam}. Points with low intensity are ignored in practice.

\subsection{Salient Point Selection and Feature Extraction}
A LiDAR scan often consists of tens of thousands of points. It is less computationally efficient to use raw point cloud matching methods such as ICP. Moreover, the raw data contains noisy measurements that can reduce the matching accuracy, \textit{e.g.}, measurements from trees alongside a road in autonomous driving. Hence it is more robust and computationally efficient to match the points clouds with features \cite{zhang2014loam,shan2018lego}.
Instead of using geometric-only features, in this paper we utilize the feature based on both geometry and intensity information. The calibrated intensity information contains the reflectivity profile of the environment that reveals the distribution of different objects. Therefore, the intensity information also helps to recognize the identical features across multiple frames. For each point $\textbf{p}_i \in \mathcal{P}$ and its intensity value $\Tilde{\eta}_i \in \Tilde{\mathcal{I}}$, we search for the nearby points $\mathcal{N}_i \in \mathcal{P}$ and calculate the local distance distribution $\sigma_{i}^{\mathcal{G}}$ and intensity distribution $\sigma_{i}^{\mathcal{I}}$ by:
\begin{equation}
\begin{aligned}
  \sigma_{i}^{\mathcal{G}} &= \frac{1}{ |\mathcal{N}_i|}\cdot \sum_{p_{j} \in \mathcal{N}_i}(|\textbf{p}_{i} - \textbf{p}_{j}|) \\
  \sigma_{i}^{\mathcal{I}} &= \frac{1}{ |\mathcal{N}_i|}\cdot \sum_{p_{j} \in \mathcal{N}_i}(|\eta_{i} - \eta_{j}|),
\end{aligned}
\end{equation}
where $|\mathcal{N}|$ is the number of points. The features are selected based on the weighted summation:
\begin{equation}
  \sigma_{i}=w^{\mathcal{G}} \cdot \sigma_{i}^{\mathcal{G}} + w^{\mathcal{I}} \cdot \sigma_{i}^{\mathcal{I}},
\end{equation}
where $w^{\mathcal{G}}$ and $w^{\mathcal{I}}$ are the weights of geometry and intensity distribution, respectively. Note that for flat surfaces such as wall, the smoothness values are small; while for corner or edge points, the smoothness values are large. Thus, for each scan, edge features $\mathcal{P}_\mathcal{E} \in \mathcal{P}$ are selected from the points with large $\sigma$, and planar features $\mathcal{P}_\mathcal{S} \in \mathcal{P}$ are selected from the points with low $\sigma$. 

\subsection{Intensity Map Building}
The intensity map $\mathcal{M}$ contains the reflectivity distribution of the surrounding environment. For most geometric-only SLAM, the map is maintained and updated via occupancy grid \cite{khan2015adaptive} or Octomap \cite{wurm2010octomap}. The 3-D space is segmented into grid cells and each cell is represented by a probability function.
Similar ideas can be used to construct and update an intensity map. Instead of a probability function, we use the intensity observation $\mathcal{I}(\eta_i|z_{1:t})$ to represent each grid cell $m_i$ \cite{khan2016modeling}. More specifically, for each observation of grid cell at time $t$, the surface reflectivity can be updated by:
\begin{equation}
  \mathcal{M}(m_i|z_{1:t})= \mathcal{M}(m_i|z_{1:t-1}) + \frac{\eta_{m_i} - \mathcal{M}(m_i|z_{1:t-1})}{n_{m_i}},
\end{equation}
where $\mathcal{M}(m_i|z_{1:t})$ is the current intensity observation and $n_{m_i}$ is the total number of observation times on cell $m_i$. To note that, if the grid contains no object, the intensity is marked as $0$ since there is no signal reflected. 

\subsection{Scan to Map Matching}
The laser odometry is the task to estimate the transformation matrix $\textbf{T} \in SE(3)$ between the current frame to the global map. The optimal pose estimation is calculated by minimizing the geometry error and intensity error.
\subsubsection{Geometry Residual}
Similar to LOAM \cite{zhang2014loam}, the geometric error is calculated by matching the current edge and planar features with the global map. It can be achieved by minimizing the point-to-edge and point-to-plane residuals. Given an edge feature $\textbf{p}_i \in \mathcal{P}_\mathcal{E}$ and the transformed point $\hat{\textbf{p}}_i = \textbf{T}\textbf{p}_i$, we can search for two nearest points $\textbf{p}^{\mathcal{E}}_1$ and $\textbf{p}^{\mathcal{E}}_2$ from the global map. The point-to-edge residual is defined by: 
\begin{equation}
  f_{\mathcal{E}}(\hat{\textbf{p}}_{i}) = \frac{(\hat{\textbf{p}}_{i} - \textbf{p}^{\mathcal{E}}_1)\times (\hat{\textbf{p}}_{i} - \textbf{p}^{\mathcal{E}}_2)}{|\textbf{p}^{\mathcal{E}}_1-\textbf{p}^{\mathcal{E}}_2|}.
\end{equation}
Similarly, given a planar feature point $\textbf{p}_i \in \mathcal{P}_\mathcal{S}$ and the transformed point $\hat{\textbf{p}}_i$, we can search for three nearest points $\textbf{p}^{\mathcal{S}}_1$, $\textbf{p}^{\mathcal{S}}_2$, and $\textbf{p}^{\mathcal{S}}_3$ from the global map. The point-to-plane residual is defined by: 
\begin{equation}
  f_{\mathcal{S}}(\hat{\textbf{p}}_k) = (\hat{\textbf{p}}_{k} - \textbf{p}^{\mathcal{S}}_1)^T \cdot \left(\frac{(\textbf{p}^{\mathcal{S}}_1 - \textbf{p}^{\mathcal{S}}_2)\times (\textbf{p}^{\mathcal{S}}_1-\textbf{p}^{\mathcal{S}}_3)}{|(\textbf{p}^{\mathcal{S}}_1 - \textbf{p}^{\mathcal{S}}_2)\times (\textbf{p}^{\mathcal{S}}_1-\textbf{p}^{\mathcal{S}}_3)|}\right).
\end{equation}

\subsubsection{Intensity Residual}
The intensity residual is calculated by matching the features with the intensity map.
It can be achieved by minimizing the intensity residual between the current point $\textbf{p}_i$ (including both edge features and planar features) and the transformed point $\hat{\textbf{p}}_i$ in the intensity map:
\begin{equation}
  f_{\mathcal{I}}(\hat{\textbf{p}}_i) =\eta_i - \mathcal{M}(\hat{\textbf{p}}_i),
\end{equation}
\begin{figure}[!b]
\begin{center}
\vspace{-8pt}
\includegraphics[width=0.59\linewidth]{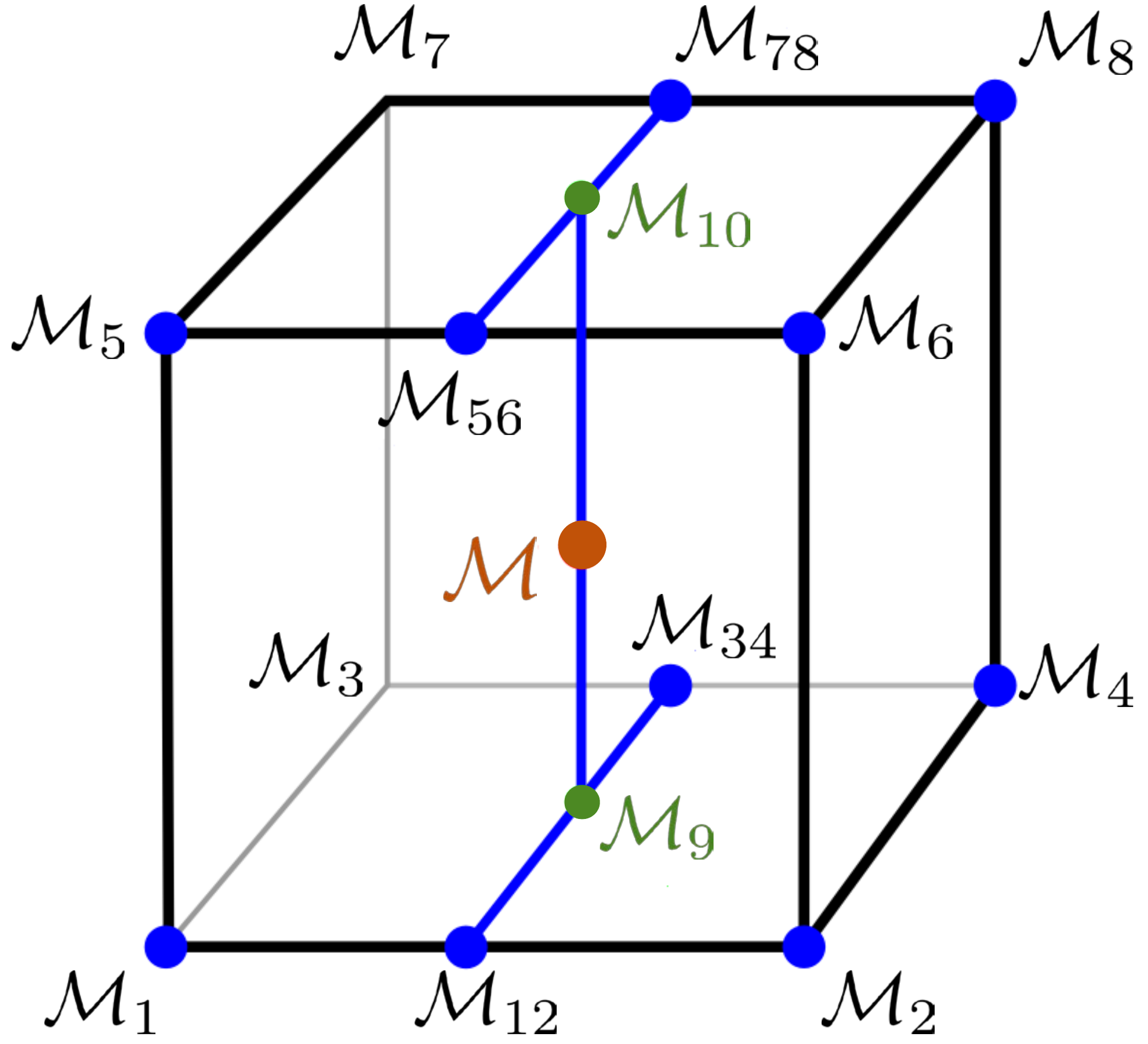}
\captionsetup{justification=justified}
\caption{Trilinear interpolation.}
\label{fig: interpolation}

\end{center}
\end{figure}
where $\mathcal{M}(\hat{\textbf{p}_i})$ is the point intensity value in the intensity map $\mathcal{M}$. To search for the intensity information in the intensity map, we introduce trilinear interpolation. Although it is more straightforward to represent the intensity by the nearest grid cell, the intensity information is less accurate especially for large scale mapping where grid resolution is low. 
For each transformed point $\hat{\textbf{p}}_i = [x_i,y_i,z_i]^T$, we can find the surrounding eight grid cells. As shown in \fref{fig: interpolation}, the intensity measurements of these grid cells are noted as $\mathcal{M}_1(\hat{\textbf{p}}_i), \mathcal{M}_2(\hat{\textbf{p}}_i), \cdots, \mathcal{M}_8(\hat{\textbf{p}}_i)$. 
Let the center position of cell 1 (closest to the origin) be $\textbf{p}_1 = [x_1, y_1, z_1]^T$ and the center position of cell 8 (farthest to the origin) be $\textbf{p}_2 = [x_2, y_2, z_2]^T$, where $x_2-x_1$, $y_2-y_1$, and $z_2-z_1$ are the width, height and depth of each grid cell, respectively. The intensity estimation of the target point is calculated as:
\begin{equation}
    \mathcal{M}_{12}(\hat{\textbf{p}}_i) = \frac{x_i - x_1}{x_2-x_1} \cdot \mathcal{M}_1(\hat{\textbf{p}}_i)+ \frac{x_i - x_1}{x_2-x_1} \cdot \mathcal{M}_2(\hat{\textbf{p}}_i),
\end{equation}
Similarly, we can derive $\mathcal{M}_{34}$, $\mathcal{M}_{56}$ and $\mathcal{M}_{78}$. The intensity value is estimated as 
\begin{equation*}
\begin{aligned}
    \mathcal{M}_{9}(\hat{\textbf{p}}_i) &= \frac{y_i - y_1}{y_2-y_1} \cdot\mathcal{M}_{12}(\hat{\textbf{p}}_i)+ \frac{y_i - y_1}{y_2-y_1} \cdot \mathcal{M}_{34}(\hat{\textbf{p}}_i) \\
    \mathcal{M}_{10}(\hat{\textbf{p}}_i) &= \frac{y_i - y_1}{y_2-y_1} \cdot\mathcal{M}_{56}(\hat{\textbf{p}}_i)+ \frac{y_i - y_1}{y_2-y_1} \cdot \mathcal{M}_{78}(\hat{\textbf{p}}_i) \\
    \mathcal{M}(\hat{\textbf{p}}_i) &= \frac{z_i - z_1}{z_2-z_1} \cdot\mathcal{M}_{9}(\hat{\textbf{p}}_i)+ \frac{z_i - z_1}{z_2-z_1} \cdot \mathcal{M}_{10}(\hat{\textbf{p}}_i). \\
\end{aligned}    
\end{equation*}

\subsubsection{Pose Estimation}
The final pose can be estimated by minimizing both geometry residual and intensity residual:
\begin{equation}
  \textbf{T}^{*} = \argmin_{\textbf{T}^{*}} \sum_{\textbf{p}_i \in \mathcal{P}_\mathcal{E}} |f_{\mathcal{E}}(\hat{\textbf{p}}_i)| + \sum_{\textbf{p}_i \in \mathcal{P}_\mathcal{S}} |f_{\mathcal{S}}(\hat{\textbf{p}}_i)| + \sum_{\textbf{p}_i \in \mathcal{P}} |f_{\mathcal{I}}(\hat{\textbf{p}}_i)| .
\end{equation}
It can be solved by the Levenberg–Marquardt algorithm \cite{more1978levenberg}.
Note that the initial pose alignment is estimated by assuming constant angular velocity and linear velocity which can increase convergence speed compared to assuming an identical transformation matrix.

\begin{figure*}[!t]
\centering
    \begin{subfigure}{0.195\linewidth}
        \begin{center}
        \includegraphics[width=0.99\linewidth]{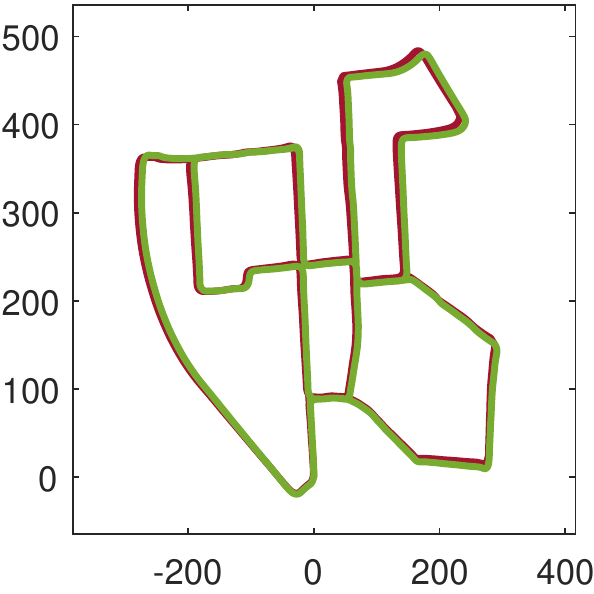}
        \end{center}
    \end{subfigure}
     \begin{subfigure}{0.195\linewidth}
        \begin{center}
        \includegraphics[width=0.99\linewidth]{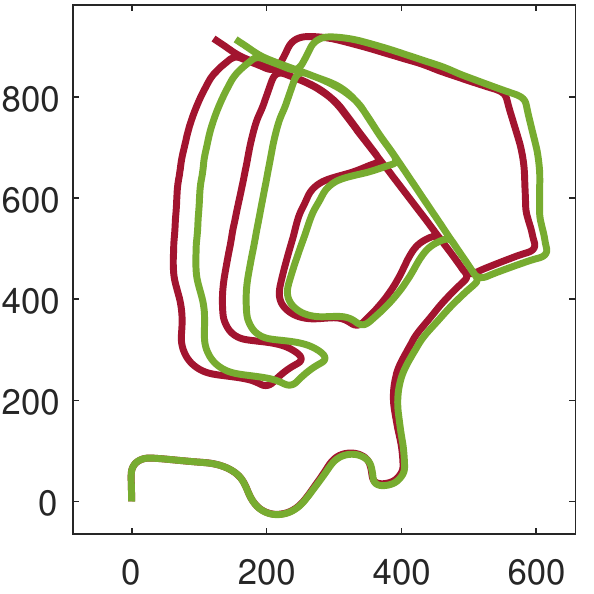}
        \end{center}
    \end{subfigure}
    \begin{subfigure}{0.195\linewidth}
        \begin{center}
        \includegraphics[width=0.99\linewidth]{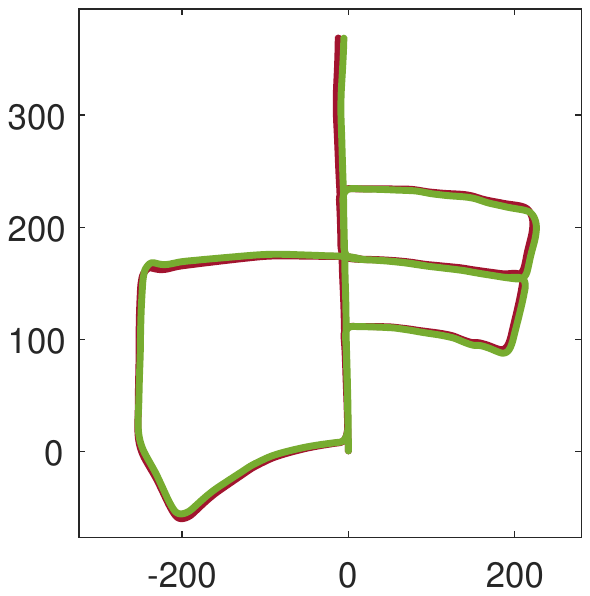}
        \end{center}
    \end{subfigure}
    \begin{subfigure}{0.195\linewidth}
        \begin{center}
        \includegraphics[width=0.99\linewidth]{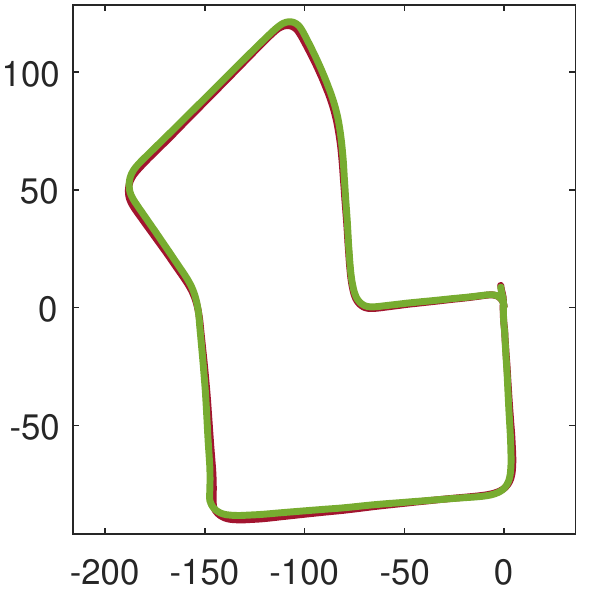}
        \end{center}
    \end{subfigure}
    \begin{subfigure}{0.195\linewidth}
        \begin{center}
        \includegraphics[width=0.99\linewidth]{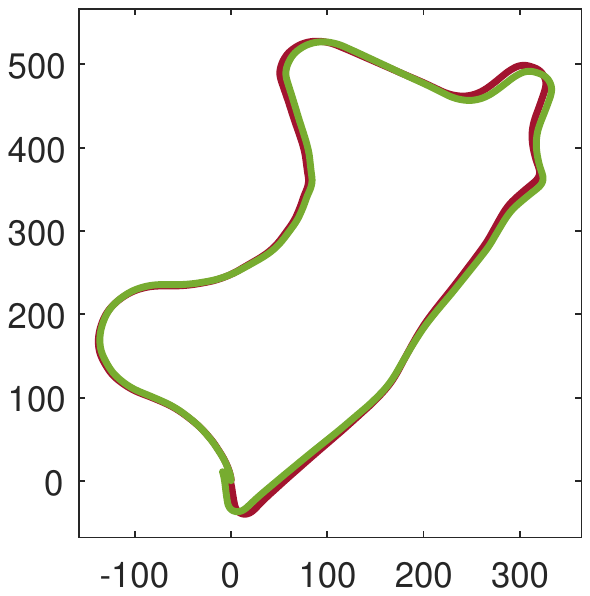}
        \end{center}
    \end{subfigure}
\caption{Result of the proposed method on KITTI dataset. From left to right are Sequence 00, 02, 05, 07, and 09. The result of intensity-SLAM is plotted in green and the ground truth is plotted in red.}
\label{fig: KITTI_result}
\end{figure*}

\subsection{Loop Closure Detection \& Global Optimization}
The goal of loop closure detection is to identify a revisited scene from historical data. For a front-end only SLAM system, it is inevitable to have drifts from the measurement noises.
To reduce the localization drifts, the back-end SLAM detects the loops by recognizing identical places. We use key frame selection to reduce the retrieval time since the computational cost of loop closure detection is often high. The key frames are selected based on the following criteria: (1) The displacement of the robot is significant, \textit{i.e.}, greater than a pre-defined threshold; (2) The change of rotation angles is significant; (3) The time elapsed is more than a certain period. For large scale environments, the thresholds are set to be higher to reduce the computational cost. All key frames are stored into a pose graph maintained in the back-end.

For each keyframe, we use Intensity Scan Context (ISC) \cite{wang2020intensity} to extract the current frame into a global signature. Compared to geometric-only descriptors such as GLAROT3D \cite{rizzini2017place}, NBLD \cite{cieslewski2016point} and Scan Context \cite{kim2018scan}, the ISC is robust to rotation change in identifying a loop closure. The ISC is a 2D matrix calculated by equally dividing polar coordinates in azimuthal and radial directions into $N_s$ sectors and $N_r$ rings. Each subspace is represented by the max intensity of the points within the area. Given a key frame, we can extract the ISC descriptor $\Omega$ from both intensity and geometry information. To compare a query ISC descriptor $\Omega^q$ with a candidate ISC descriptor $\Omega^c$, let $\mathbf{v}_i^q$ and  $\mathbf{v}_i^c$ be the $i^{th}$ column of $\Omega^q$ and $\Omega^c$. The similarity score $\varphi(\Omega^q,\Omega^c)$ is found by taking the mean cosine distance of each sector:
\begin{equation}
 \begin{aligned}
  \varphi(\Omega^q,\Omega^c) = \frac{1}{N_s} \sum_{i=0}^{N_s-1}(\frac{{\mathbf{v}_i^{q}}^T \cdot \mathbf{v}_i^c}{\norm{\mathbf{v}_i^q}\cdot \norm{\mathbf{v}_i^c}}).
\end{aligned}
\end{equation}
The loop closure can be determined by setting a threshold on $\varphi(\Omega^q,\Omega^c)$. 

Loop closure detection is able to efficiently identify loop pairs. However, false detection can lead to failure of pose graph optimization. In order to prevent false positives, geometry consistency verification is used to check the similarity of candidate frames. For the candidate loop frame, we search for the nearby line and plane information from the global map. And for current frame, edge and planar features are extracted and matched to the corresponding global lines and planes by minimizing the point-to-edge and point-to-plane distances. When two frames are unrelated, the sum of distances is often high. Hence the false positive can be filtered out by setting a threshold on it. With the revisited place identified, we can add the edge between two frames into the pose graph and global optimization can be applied to correct the drifts \cite{grisetti2010tutorial}.

%% file: body/Experiments.tex
\subsection{Experiment Setup}
In this section we present comprehensive experiments of the proposed Intensity-SLAM, including the evaluation on public dataset and practical experiments on a real robot platform.
For public dataset test, the proposed method is evaluated in KITTI \cite{Geiger2012CVPR} that is commonly used for SLAM evaluation.
The algorithm is implemented in C++ on Robotics Operating System (ROS) Melodic and Linux Ubuntu 18.04. In order to measure the computational cost of different approaches, all algorithms are tested on a desktop computer equipped with an Intel 6-core i7-8700 processor CPU. The instruction set of Eigen is SSE2. In the practical test, the proposed method is integrated into an Automated Guided Vehicle (AGV) in a warehouse environment. In the AGV test, our algorithm is running in an Intel NUC mini PC with an Intel i7-10710U processor which is often used on mobile robots. 


\subsection{Evaluation Metric}
To evaluate the accuracy of localization, We use Average Translational Error (ATE) and Average Rotational Error (ARE) \cite{Geiger2012CVPR}: 
\begin{equation}
\begin{aligned}
  E_{rot}(\mathcal{F}) &= \frac{1}{|\mathcal{F}|} \sum_{i,j \in \mathcal{F}} \angle[\hat{\textbf{T}}_j \hat{\textbf{T}}_i^{-1} \textbf{T}_i \textbf{T}_j^{-1} ]\\
  E_{trans}(\mathcal{F}) &= \frac{1}{|\mathcal{F}|} \sum_{i,j \in \mathcal{F}} {||\hat{\textbf{T}}_j \hat{\textbf{T}}_i^{-1} \textbf{T}_i \textbf{T}_j^{-1} ||}_2,\\
\end{aligned}
\end{equation}
where $\mathcal{F}$ is a set of frames $(i, j)$, $\textbf{T}$ and $\hat{\textbf{T}}$ are estimated and true LiDAR poses respectively, and $\angle[\cdot]$ is the rotation angle. The computing cost is measured by the average processing time of each frame.

\subsection{Evaluation on Public Dataset}
The KITTI dataset is collected from an autonomous driving car running in a large scale urban area. It is challenging due to the complex environment with both static and dynamic objects. The proposed method is firstly evaluated on the KITTI dataset with various scenarios such as urban, highway and country. In total we tested over more than 10,000 frames with more than 10 km travelling distance. Some of the experimental results are shown in \fref{fig: KITTI_result}, in which Sequence 00, 02, 05, 07, and 09 are selected for demonstration. The trajectory generated from the proposed method is in green and the ground truth is in red. It can be seen that our method is able to track the robot's pose accurately.

\subsection{Ablation Study of Intensity Information}
\begin{figure}[t]
\begin{center}
\vspace{8pt}
\includegraphics[width=0.89\linewidth]{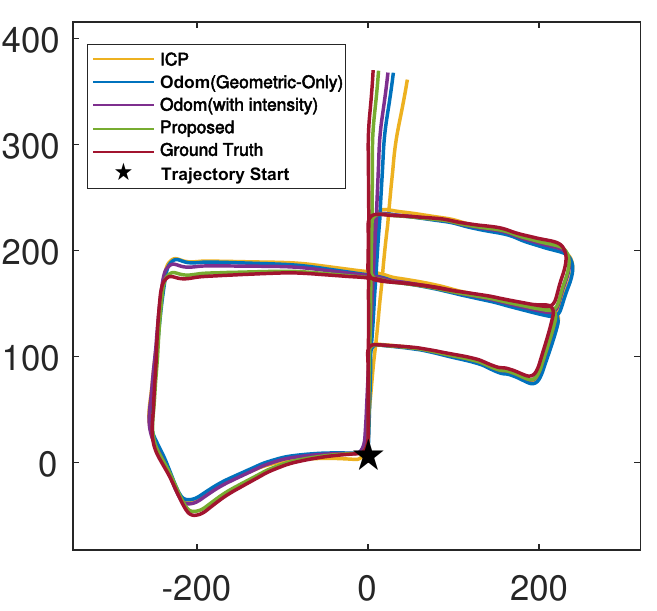}
\captionsetup{justification=justified}
\caption{Performance comparison on KITTI Sequence 05.}
\label{fig: comparison}
\end{center}
\end{figure}

We also demonstrate the effect of intensity information in localization. To be consistent, the loop closure detection is removed and only front-end laser odometry is compared. In the experiment, the ``geometric-only laser odometry" method only includes point-to-edge and point-to-plane residual, while ``odometry with intensity" method integrates intensity residual into the geometric-only laser odometry. All other parameters are kept the same. Moreover, we also include the result of the Iterative Closest Point method for comparison. The results are shown in \fref{fig: comparison} and \tref{table:ablation}. 
In particular, we use KITTI Sequence 05 for illustration. It can be seen that by adding the intensity residual, the overall ATE is reduced from $0.69\%$ to $0.62\%$ and the overall ARE is reduced from $5.7\times10^{-5} \deg/m$ to $5.3\times10^{-5}\deg/m$. The intensity information improves the translational and rotational accuracy by around 10\%, while it does not increase the computing time too much. 

\subsection{Comparison with Existing Methods}
We also compare our approach with state-of-the-art methods including LeGO-LOAM \cite{shan2018lego}, A-LOAM, and HDL-GRAPH-SLAM \cite{koide2018portable}. 
All approaches are tested using the same settings and the results are shown in \tref{table: comparison}. It can be seen from the table that our method is competitive in terms of computational cost. 
And it is worthy noting that our method also achieves the best accuracy and outperforms existing geometric-only methods.


\begin{table}[t]
    \begin{center}
    \begin{tabular}{cccc}
    \toprule
    Method & Time (ms) & ATE (\%) & ARE (deg/m)  \\
    \midrule
    Iterative Closest Point  &142 & 1.031 &   0.0092    \\ 
    Odometry(geometric-only)  &\textbf{92} & 0.691 &   0.0057    \\ 
    Odometry(with intensity)  &113 & 0.627 &   0.0053   \\  
    \textbf{Intensity-SLAM}  &115 &\textbf{ 0.529 }    &   \textbf{0.0044} \\  
    \bottomrule
    \end{tabular}
    \caption{Result analysis of intensity-aided method and non-intensity-aided method on KITTI the dataset. }
    \label{table:ablation}
    \end{center}
\end{table}

\subsection{Validation on Practical Robots}
\begin{figure}[t]
\begin{center}
\vspace{8pt}
\includegraphics[width=0.99\linewidth]{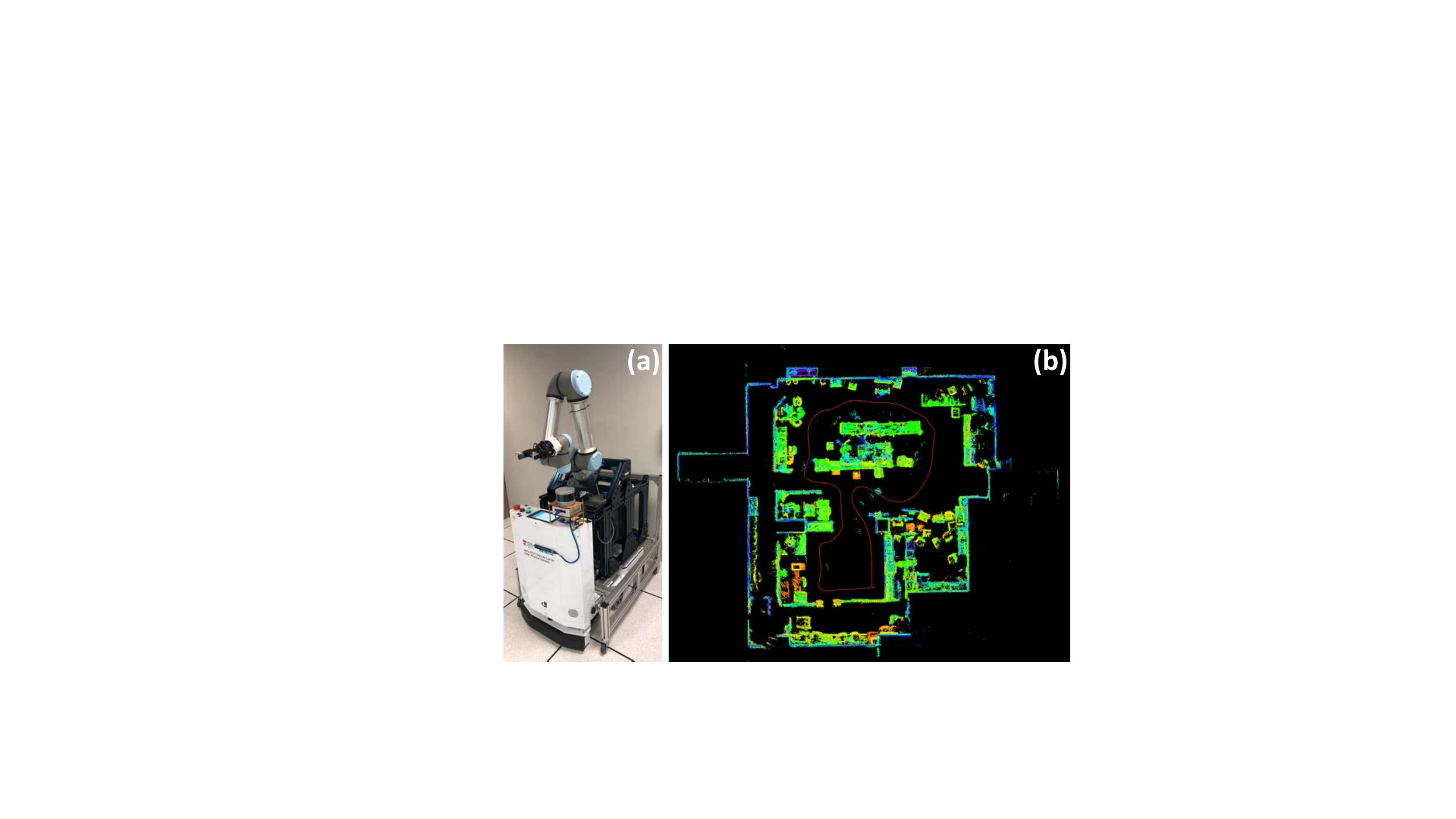}
\captionsetup{justification=justified}
\caption{Localization and mapping result in warehouse environment. (a) AGV platform for warehouse manipulation. (b) Localization and mapping result of the proposed method. The robot's trajectory is plotted in red color.}
\label{fig: warehouse}
\end{center}
\end{figure}

In this section, we test the performance of our method on a practical robot in a warehouse environment.
To this end, we implement our algorithm on an industrial AGV that is used for grabbing and transporting materials. As shown in \fref{fig: warehouse} (a), the AGV is equipped with an Intel NUC mini computer and a Velodyne VLP-16 LiDAR for perception and localization. In this experiment, the robot uses Intensity-SLAM for localization and is supposed to fetch items from the target storage shelf and return. The localization and mapping result is shown in \fref{fig: warehouse} (b), with the robot's trajectory plotted in red. It can be seen that our method is able to provide reliable localization and generate a dense 3D map for a complex indoor environment. In particular, the robot is manually controlled by a joystick to return to the same location and we record the final localization drift. The final localization drift is about 5 \textit{cm}, which is accurate enough for general indoor robotic applications.

\begin{table}[t]
    \begin{center}
    \begin{tabular}{cccc}
    \toprule
    Methods & Time (ms) & ATE (\%) & ARE (deg/m)  \\
    \midrule
    \textbf{Intensity-SLAM}   &115 &\textbf{ 0.529 }    &   \textbf{0.0044} \\  
    LeGO-LOAM &\textbf{84} &0.724    &   0.0072 \\  
    HDL-Graph-SLAM & 346&1.127    &   0.0098 \\  
    A-LOAM & 182 & 0.702    &   0.0058 \\  
    \bottomrule
    \end{tabular}
    \caption{Performance comparison of different approaches on KITTI Sequence 05.}
    \label{table: comparison}
    \end{center}
\end{table}

%% file: body/Conclusion.tex
In this paper, we present a novel intensity-assisted full SLAM framework for LiDAR-based localization system. Existing methods mainly leverage geometric-only features to estimate location and ignore intensity information. We begin with the physical model of intensity measurement and illustrate the use of intensity information. We introduce the intensity map and intensity residual for pose estimation to improve the localization accuracy. Moreover, we propose a full SLAM structure that consists of both front-end and back-end systems, namely Intensity-SLAM. Thorough experiments are performed in different scenarios, including indoor AGV mapping in warehouse and outdoor autonomous driving in urban areas. The experimental results show that by integrating intensity information, the localization accuracy is significantly improved. The proposed method also achieves much lower translational error and rotational error compared to the state-of-the-art geometric-only methods. Our method is publicly available. 